\documentclass{article}
\usepackage{22}

\usepackage{times}
\usepackage{soul}
\usepackage{url}
\usepackage[hidelinks]{hyperref}
\usepackage[utf8]{inputenc}
\usepackage[small]{caption}
\usepackage{graphicx}
\usepackage{amsmath}
\usepackage{amsthm}
\usepackage{booktabs}
\usepackage{algorithm}
\usepackage{algorithmic}
\usepackage{subfigure}

\usepackage[utf8]{inputenc}
\usepackage{adjustbox}
\usepackage{amsmath, amsfonts}
\title{Exploiting Semantic Epsilon Greedy Exploration Strategy in \\ Multi-Agent Reinforcement Learning}

\author{
Hon Tik Tse, Ho-fung Leung
\affiliations
The Chinese University of Hong Kong
\emails
1155126684@link.cuhk.edu.hk, lhf@cuhk.edu.hk
}

\begin{document}

\maketitle

\begin{abstract}
Multi-agent reinforcement learning (MARL) can model many real-world applications. However, many MARL approaches rely on $\varepsilon$-greedy for exploration, which may discourage visiting advantageous states in hard scenarios. In this paper, we propose a new approach QMIX(SEG) for tackling MARL. It makes use of the value function factorization method QMIX to train per-agent policies and a novel \textbf{S}emantic \textbf{E}psilon \textbf{G}reedy (SEG) exploration strategy. SEG is a simple extension to the conventional $\varepsilon$-greedy exploration strategy, yet it is experimentally shown to greatly improve the performance of MARL. 
We first cluster actions into groups of actions with similar effects and then use the groups in a bi-level $\varepsilon$-greedy exploration hierarchy for action selection. We argue that SEG facilitates semantic exploration by exploring in the space of groups of actions, which have richer semantic meanings than atomic actions. Experiments show that QMIX(SEG) largely outperforms QMIX and leads to strong performance competitive with current state-of-the-art MARL approaches on the StarCraft Multi-Agent Challenge (SMAC).

\end{abstract}

\section{Introduction}

Multi-agent systems  (MASs) can be used to model many complex scenarios. Multi-agent reinforcement learning (MARL) has been growing in popularity and importance, and it is used in many real-world applications, such as 
robot swarm control~\cite{huttenrauch}.

Different MARL algorithms have been designed to enable multiple agents to learn to make good decisions.
Among them, centralized training with decentralized execution (CTDE) has been shown to be a powerful paradigm achieving better performance than algorithms with non-centralized training~\cite{gupta17,coma}. 
An influential algorithm in the CTDE paradigm is QMIX~\cite{qmix}, which 
estimates joint action-values as a  non-linear monotonic function of per-agent values.
A lot of work~\cite{qtran,wqmix,pmlr-v119-wang20f,refil,qplex,rode} has been put into building on QMIX to develop better-performing algorithms. Recently, RODE~\cite{rode}, a hierarchical algorithm utilizing QMIX, has seen huge success in the StarCraft Multi-Agent Challenge~\cite{smac} (SMAC). 
The novelty of RODE is twofold.  First, the action space is decomposed into \emph{roles} consisting of actions of similar functionalities. Second, the action spaces of the agents are limited by the roles they select, that is, agents must select actions from the restricted action spaces corresponding to the roles they selected.

The restriction of action spaces in RODE improves learning, but places heavy limitations on the actions agents can select. In the RODE framework, agents are required to follow a selected role for extended time steps. This reduces the flexibility of an agent and prevents the agent from reacting to rapid changes in the environment. 

In fact, as 
Nachum~{\it et~al.}~\shortcite{whydoes} point out, in the single-agent case, hierarchical reinforcement learning algorithms attain better performance mainly due to two key improvements in exploration: \textit{temporally extended exploration} and \textit{semantic exploration}. 
They further show that these exploration benefits can actually be recreated in non-hierarchical agents employing carefully designed exploration strategies. 
We note that Dabney~{\it et~al.}~\shortcite{temporalEpsilon} have demonstrated in a recent paper  
that the \emph{temporally extended $\varepsilon$-greedy exploration}, a simple extension of $\varepsilon$-greedy exploration, 
can improve the performance of reinforcement learning in hard-exploration Atari games with minimal loss in performance on the remaining games.

In this paper, we propose a novel approach QMIX(SEG) for cooperative MARL scenarios. QMIX(SEG) is a non-hierarchical learning approach. It first employs QMIX to represent the Q-values of joint actions as a nonlinear monotonic function of per-agent Q-values of actions. Then, it uses a novel \textbf{S}emantic \textbf{E}psilon \textbf{G}reedy (SEG) exploration strategy for action selection.
SEG is a simple yet effective 2-level extension to $\varepsilon$-greedy strategy achieving semantic exploration. 
Conceptually, at the higher level, SEG explores in the space of groups of actions, while at the lower level it explores in the space of atomic actions within the selected group.

Intuitively, QMIX(SEG) aims at recreating the exploration benefits exhibited in the hierarchical learning framework RODE in a non-hierarchical framework.  Meanwhile, it lifts the restrictions RODE places on the actions agents can select, that is, they can only be selected from the currently selected role for multiple steps. QMIX(SEG) is also an attempt to extend $\varepsilon$-greedy exploration to develop a still simple, yet effective exploration method that removes the inherent bias in $\varepsilon$-greedy. 

Experimental results on SMAC show that QMIX(SEG) outperforms other state-of-the-art approaches, including QMIX, RODE, MAPPO(FP), and MAPPO(AS)~\cite{mappo}. In particular, QMIX(SEG) can attain the highest or near-highest win rates on 4 out of 5 super hard maps. QMIX(SEG) also achieves strong results competitive with DDN~\cite{dfac}, a much more complex approach to generalize expected value function factorization methods. 

\section{Related Work}

In recent years, there has been increasing interest in improving multi-agent reinforcement learning (MARL) algorithms by innovative exploration mechanisms. Wang~{\it et~al.}~\shortcite{influenceExp} exploit interaction between agents for exploration while Jaques~{\it et~al.}~\shortcite{socialInfluence} propose to reward actions that have a large influence on other agents' behaviors. Mahajan~{\it et~al.}~\shortcite{maven} explore the use of a shared latent variable to coordinate exploration and encourage diverse behavior by maximizing the mutual information between the trajectories and the latent variable. Liu~{\it et~al.}~\shortcite{cmae} encourage agents to explore the projected state space while increasing the dimension of the projected state space gradually. Gupta~{\it et~al.}~\shortcite{uneven} propose to perform actions of similar solved tasks on the target task. 

Hierarchical reinforcement learning (HRL)~\cite{SUTTON1999181} has been shown to have superior performance, especially in sparse reward problems~\cite{10.1145/3453160}. It has also been extensively studied in the context of MARL~\cite{doi:10.1080/01969722.2019.1677335,pmlr-v119-wang20f,Lee2020Learning,10.5555/3398761.3398941}.

In a recent work, Nachum~{\it et~al.}~\shortcite{whydoes}
show that the benefits of HRL mainly come from improved explorations: in HRL, there are both \emph{temporally extended explorations} of the environment across multiple environment steps, and \emph{semantic explorations} of semantically meaningful actions. They further show that such benefits can be achieved in non-hierarchical agents with certain exploration methods, allowing non-hierarchical agents to achieve competitive results with hierarchical agents. 

Dabney~{\it et~al.}~\shortcite{temporalEpsilon} propose to extend $\varepsilon$-greedy exploration, achieving results comparable to more complex non-dithering exploration methods. They focus on extending $\varepsilon$-greedy exploration in the direction of temporally extended exploration by repeating a selected action for a random number of steps.

RODE~\cite{rode} is a hierarchical MARL algorithm that demonstrates superiority in performance over existing algorithms in scenarios of SMAC. An innovative idea of RODE is that each agent first selects a \emph{role} using $\varepsilon$-greedy exploration, and then an action available in the role action space, once again using $\varepsilon$-greedy exploration. 

\section{Methods}

\begin{figure*}
    \centering
    \begin{minipage}[]{0.65\textwidth}
        \centering
        \includegraphics[width=0.75\textwidth]{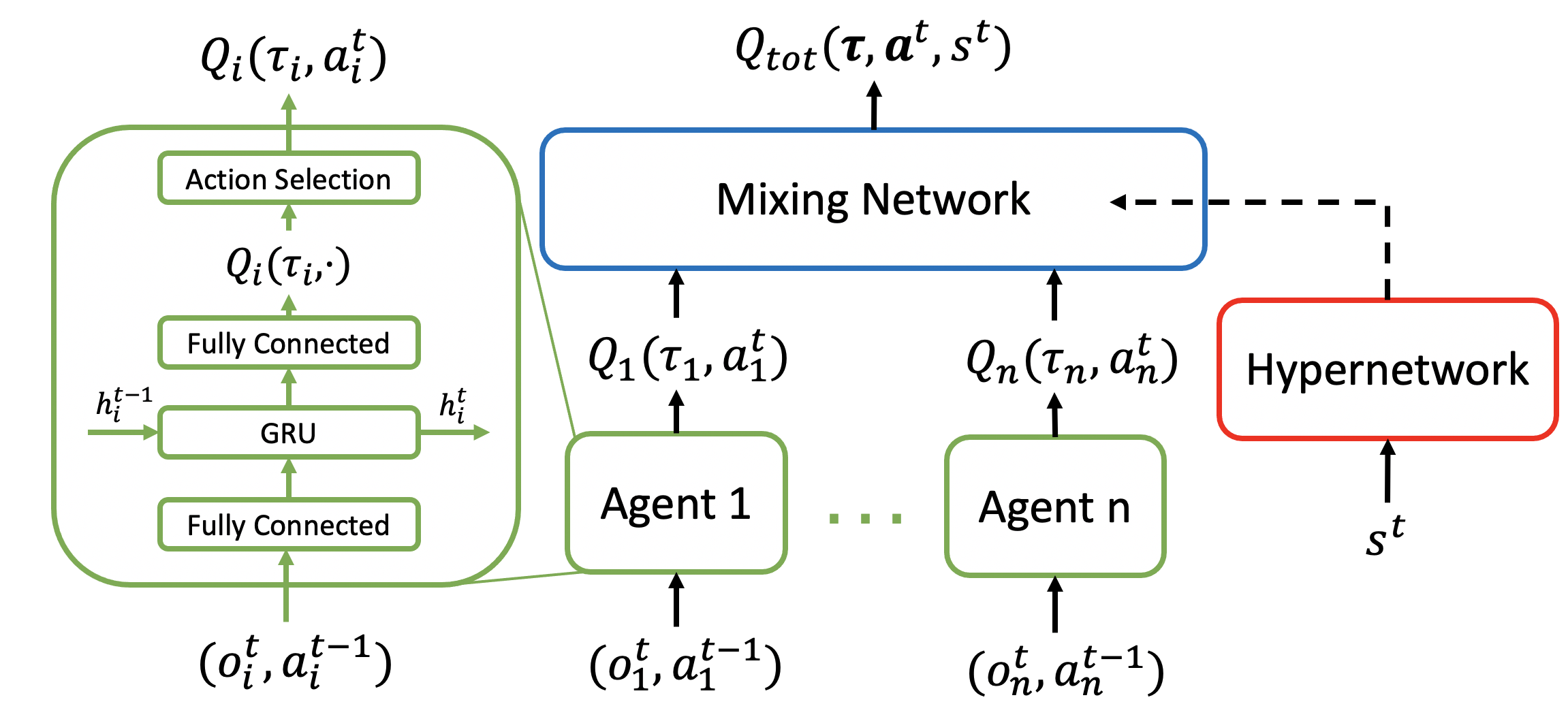}
        \caption{Q-network and mixing network same as in QMIX.}
        \label{fig:QNetwork}
    \end{minipage}\hfill
    \begin{minipage}{0.3\textwidth}
        \centering
        \includegraphics[width=0.75\textwidth]{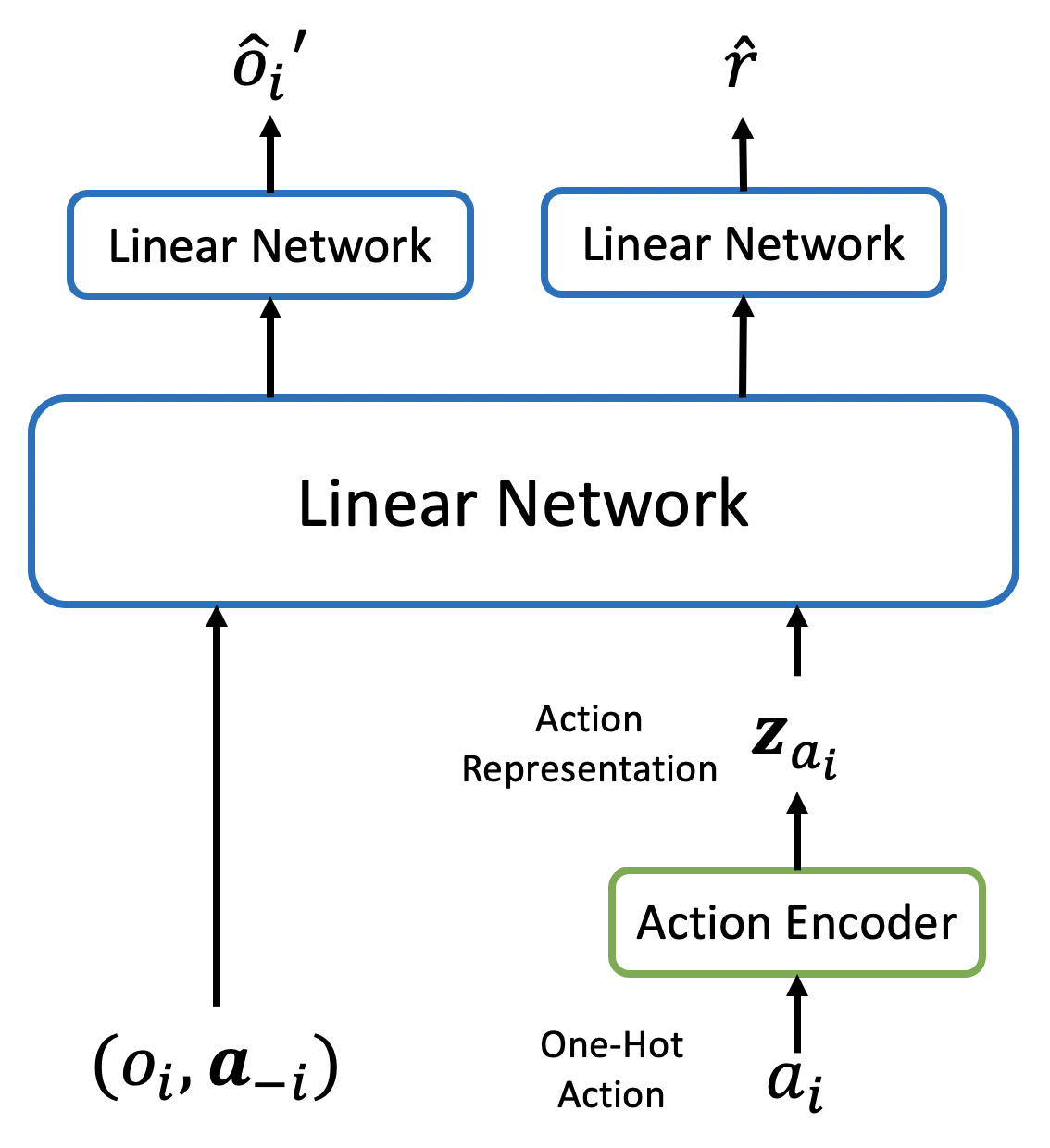} 
        \caption{Forward model for learning action representations same as in RODE.}
        \label{fig:ActionRepresentation}
    \end{minipage}
\end{figure*}

In this section, we propose a novel approach QMIX(SEG) for tackling cooperative multi-agent problems. The novelty comes from an exploration strategy called \textbf{S}emantic \textbf{E}psilon \textbf{G}reedy (SEG), which adds an extra layer of $\varepsilon$-greedy exploration to the conventional $\varepsilon$-greedy exploration. Our method is inspired by RODE, and it extends $\varepsilon$-greedy exploration in the direction of semantic exploration.

\subsection{Background}

A fully cooperative multi-agent task can be formulated as a decentralized partially observable Markov decision process (Dec-POMDP)~\cite{Dec-POMDP} $\langle S,A,P,R,\Omega,O,n,\gamma \rangle$, where $n$ is the number of agents,  $\gamma \in [0,1)$ is the discount factor and $s\in S$ is the true state of the environment. At each time step, each agent $i$ takes an action $a_i \in A$,  causing the environment to move into the next state $s'\in S$ and $s' \sim P(s'\mid s,\boldsymbol{a})$ where $\boldsymbol{a} =(a_1,\dots a_n)\in A^n$ is the joint action. Agents also receive a shared reward $r=R(s,\boldsymbol{a})$. We consider a partially observable setting so each agent $i$ receives an observation $o_i \in \Omega$ and $o_i=O(s,i)$. Each agent has an action-observation history $\tau_i \in T\equiv (\Omega,A)^*$, on which it conditions its policy $\pi_i(a_i|\tau_i)$. The joint policy $\boldsymbol{\pi}=(\pi_1,\dots \pi_n)$ has a joint action-value function $Q^{\boldsymbol{\pi}}_{tot}(s,\boldsymbol{a})= \mathbb{E}_{s_{0:\infty},\boldsymbol{a}_{0:\infty}}[\sum_{t=0}^\infty \gamma^t r_t\mid s_0=s,\boldsymbol{a}_0=\boldsymbol{a},\boldsymbol{\pi}]$.

In the centralized training with decentralized execution (CTDE) paradigm, the learning algorithm has access to the true state $s$ of the environment and the action-observation histories  $\boldsymbol{\tau}=(\tau_1,\dots \tau_n)$ of all agents during training. On the other hand, each agent's policy $\pi_i$   depends only on its local action-observation history $\tau_i$, allowing for decentralized execution.
A popular approach to achieve CTDE is \emph{value function factorization}, which decomposes the joint Q-function into per-agent utility functions, for example as their sum~\cite{vdn}. QMIX~\cite{qmix}  factorizes the joint Q-function into a non-linear monotonic function of per-agent utility functions, and its success triggers several subsequent research~\cite{qtran,wqmix,qplex}.

\subsection{Learning Q-values Using QMIX}
Before going into how QMIX(SEG) makes use of Q-values for exploration, we first describe how Q-values are learned. We follow exactly the approach in QMIX~\cite{qmix} and use a Q-network to estimate the per-agent utility functions and a mixing network to estimate the joint Q-function. We use value function factorization as it allows agents to better learn coordinated behavior. We use a QMIX-style mixing network for the monotonic factorization it provides and its popularity. The mixing network is only used during centralized training, and is not used during decentralized execution. To stay self-contained, we describe the Q-network, the mixing network, and the process of learning their parameters in this subsection. Figure~\ref{fig:QNetwork} illustrates the Q-network and the mixing network. 

The Q-network consists of a fully connected layer, followed by a GRU, and further followed by a fully connected layer. The Q-network takes the local observation and the last local action selected as input. The GRU allows the agents to utilize past actions and observations for predicting the Q-values of actions, which are the outputs of the last fully connected layer. We utilize parameter sharing and agents share the same Q-network.

Suppose each agent $i$ selects the action $a_i$. The Q-values of the selected actions $Q_1(\tau_1,a_1),\dots,Q_n(\tau_n,a_n)$ are passed as input to the mixing network, which outputs the joint Q-value $Q_{tot}(\boldsymbol{\tau},\boldsymbol{a},s)$. The parameters of the mixing network are predicted by a hypernetwork~\cite{HaDL17} taking the state $s$ as input. In particular, the weights of the mixing network predicted by the hypernetwork are non-negative such that the joint Q-function is a monotonic function of per-agent utility functions. The biases, on the other hand, do not need to be non-negative. Passing the state $s$ as input to the hypernetwork allows the joint Q-function to depend on the state $s$ in non-monotonic ways. The Q-network, parameterized by $\theta$, and the mixing network, parameterized by $\phi$, are trained end-to-end by minimizing the following loss function:
\[
\mathcal{L}(\theta,\phi)=\mathbb{E}_{\mathcal{D}}[(r+\gamma \max_{\boldsymbol{a'}} Q_{tot}^-(\boldsymbol{\tau'},\boldsymbol{a'},s')-Q_{tot}(\boldsymbol{\tau},\boldsymbol{a},s))^2].
\]
Here, $Q^-_{tot}$ is a target network, $\boldsymbol{\tau'}$ is the next joint action-observation history, $s'$ is the next state, and $\mathcal{D}$ is a replay buffer from which uniform samples are drawn. The maximization of $Q^-_{tot}$ in the loss function can be easily done.  Since the joint Q-function is a monotonic function of per-agent utility functions, maximizing per-agent utility functions then passing the max values and the state into the mixing network gives $\max Q_{tot}^-$. After the Q-values are learnt, we use SEG for action selection and exploration.

\subsection{Semantic Epsilon Greedy}

In the Semantic Epsilon Greedy (SEG) exploration strategy, we first learn to cluster actions into groups of actions with similar effects. Then, we adopt a bi-level $\varepsilon$-greedy hierarchy to ensure each group of actions is explored with equal probability. We use "semantic" instead of "hierarchical" as the name of the exploration strategy to avoid confusion between our method, a hierarchical {\it exploration strategy}, and hierarchical {\it learning algorithms}.

The approach in SEG to cluster actions into groups is the same as that of RODE. We identify the groups of actions by first learning the action representations of all actions and then clustering the actions into groups by their representations. 

To learn the action representations that reflect the effect of actions on the environment and other agents, we use the induced reward and the change in local observations to measure the effect of an action. Figure~\ref{fig:ActionRepresentation} illustrates the forward model for learning action representations. First, we encode the one-hot action $a_i$ that agent $i$ takes as a $d$-dimensional representation $\boldsymbol{z}_{a_i}$ using an action encoder $f_e(\cdot;\theta_e)$. 
Then, the action representation $\boldsymbol{z}_{a_i}$ is passed as input to the observation predictor $p_o$ and reward predictor $p_r$ to predict the next local observation and the received reward respectively, given the current local observation $o_i$ of agent $i$ and the one-hot actions $\boldsymbol{a}_{-i}$ of all other agents. The intuition is that if the representation can be used to predict the next local observation and reward, the representation captures the effect of the action on the environment and other agents.
The action encoder $f_e$, parameterized by $\theta_e$, and the observation predictor $p_o$ and reward predictor $p_r$, jointly parameterized by $\xi_e$, are trained end-to-end by minimizing the following loss function:
\[
\mathcal{L}_e(\theta_e,\xi_e)=\mathbb{E}_{(\boldsymbol{o},\boldsymbol{a},r,\boldsymbol{o}')\sim \mathcal{D}}[\sum_{i=1}^n \| \hat o'_i-o'_i\|^2_2+\lambda_e \sum_{i=1}^n (\hat r -r)^2],
\]
where  $\mathcal{D}$ is the replay buffer, $\hat o'_i=p_o(\boldsymbol{z}_{a_i},o_i,\boldsymbol{a}_{-i})$ is the predicted next local observation of agent $i$, $\hat r=p_r(\boldsymbol{z}_{a_i},o_i,\boldsymbol{a}_{-i})$ is the predicted received reward,  $o'_i$ is the true next local observation of agent $i$, $r$ is the true received reward, and the sum is carried out over all agents. The scaling factor $\lambda_e$ balances between the error of predicting next local observations and the error of predicting the received reward. The trained $f_e$ is then used to encode all actions into their corresponding action representations, which are fixed for the remaining time steps. 

After obtaining the action representations, we cluster the actions into groups of actions using $k$-means clustering based on Euclidean distances between action representations. The number of clusters or groups is chosen as a hyperparameter. Again we follow RODE and use Euclidean distances as the distance metric. Actions belonging to the same group are generally similar in effect (for example, they are all {\it attack} actions of some kind), as they have small Euclidean distances among their action representations.

After clustering the actions into groups of actions, we further add an action that is always available to all agents called {\it no-op} (no operation) to each group. The existence, name, and number of such actions may vary in different benchmarks. For example, in SMAC, such actions are called {\it no-op} and {\it stop}.  This is to ensure that in each group of actions, at least one action is always available to all agents. Note that domain knowledge, if available, can be used to directly cluster the actions into groups without having to learn and cluster action representations.

After clustering the actions into groups, SEG makes use of the groups for bi-level exploration. The following hierarchical exploration procedure is carried out for each agent $i$ at each time step during training. Let $Q_i(\tau_i,a)$ be the Q-value of action $a\in A$ for agent $i$. 
\begin{enumerate}
\item At the higher level, with probability $1-\varepsilon$, we select the \emph{action} $\arg\max_{a\in A}Q_i(\tau_i,a)$ with the largest Q-value, just as in standard $\varepsilon$-greedy exploration.
\item However, the difference from standard $\varepsilon$-greedy exploration lies in what we do with probability $\varepsilon$. Specifically, with probability $\varepsilon$,
\begin{enumerate}
\item we select a \emph{group} $A^j$ uniformly randomly, and  
\item after selecting the group $A^j$, we carry out the lower level of $\varepsilon$-greedy exploration of actions:
with probability $1 - \varepsilon$, we select the action $\arg\max_{a\in A^j}Q_i(\tau_i,a)$ with the largest Q-value that \emph{is available in the selected group $A^j$}; with probability $\varepsilon$, we select one of the actions \emph{available in the selected group $A^j$} uniformly randomly. 
\end{enumerate}
\end{enumerate}

This bi-level procedure is carried out for each agent, meaning that different agents, when they explore simultaneously, may explore in different groups.

SEG is a bi-level extension of $\varepsilon$-greedy. At the higher level, SEG can be interpreted as exploring the space of groups of actions. Suppose the number of groups is $m$, then the actions can be clustered into $A^1,\dots,A^m$. The higher level of SEG can be interpreted as selecting from the following groups: $A, A^1, \dots,A^m$, where the set of all available actions $A$ is also interpreted as a group. If we associate a Q-value with each of $A,A^1,\dots,A^m$ and assume the set of all actions $A$ always has the highest Q-value, then using conventional $\varepsilon$-greedy to select one of $A,A^1,\dots,A^m$ is very similar to what we do in SEG, which is to select the group with the highest Q-value $A$ with probability $1-\varepsilon$ and select one of the remaining groups $A^1,\dots,A^m$ uniformly randomly with probability $\varepsilon$.

SEG achieves semantic exploration by exploring at the higher level the space of groups of actions, which has richer semantic meanings than atomic actions. The intuitive reason semantic exploration helps is that by exploring in the space of groups where elements have richer semantic meanings, the exploration carried out will have richer meaning in the environment. 
A good example proposed by Nachum~{\it et~al.}~\shortcite{whydoes} is that in a robot navigation task, exploring at the level of $x$-$y$ coordinates is more sensible than exploring at the level of robot joint torques.  

Note that SEG only limits the actions agents can select when agents explore and select groups (with probability $\varepsilon$) during training, in which case the actions must be selected from the selected groups. SEG does not limit the actions agents can select during test time as $\varepsilon=0$ during test time.

As long as clustering of actions into groups of similar actions is present, SEG can be implemented easily. Classifying actions into groups of similar actions can be done easily for discrete actions, so SEG can be implemented easily for problems with discrete action space. SEG is an exploration method that only requires the Q-values of actions and clustering of actions, allowing it to be applied generally.
In fact, SEG does not exploit centralized training in any way, making SEG compatible with algorithms even not in the CTDE paradigm, such as IQL~\cite{iql}.

\subsection{Discussion}

A more practical way of looking at why SEG gives rise to better exploration than standard $\varepsilon$-greedy exploration is to look at the probability of exploring states that require coordination for extended time steps to reach. Suppose there exist states that require all $n$ agents to take actions with similar effects (or, actions in the same group) for $t$ steps to reach. Denote $s^i$ as the state that requires all $n$ agents to take actions in the group $A^i$ for $t$ steps to reach. The probability of reaching $s^i$ with standard $\varepsilon$-greedy exploration is $(\frac{\varepsilon |A^i|}{|A|})^{nt}$. Note that the probability depends on the number of actions in the group $A^i$. This means that standard $\varepsilon$-greedy is biased towards visiting states that can be reached by a large number of similar actions. This problem is present in single-agent reinforcement learning but is worsened in MARL because of the extra $n$ in the exponent. This is undesirable if states that are critical to good performance can only be reached by jointly selecting a small number of similar actions.

SEG removes this bias in standard $\varepsilon$-greedy by exploring each group with equal probability. Since SEG first selects a group uniformly randomly before selecting an atomic action from the selected group, suppose the number of groups is $m$, the probability of reaching $s^i$ for all $i$ is $(\frac{\varepsilon}{m})^{nt}.$ Now, the probability of reaching $s^i$ is independent of the number of actions in the group $A^i$ required to reach $s^i$. SEG, by removing the bias in standard $\varepsilon$-greedy, allows the agents to visit $s^i$'s with equal probability. In simpler settings where jointly selecting groups containing a large number of actions already give rise to optimal behavior, SEG may result in slower learning due to excessive exploration. However, in harder scenarios where advantageous states can only be reached by a small number of similar actions, SEG allows the agents to visit these states more and learn better strategy while using $\varepsilon$-greedy for exploration may fail to reach these states. This provides a more practical foundation for why SEG can achieve superior performance compared to standard $\varepsilon$-greedy exploration, especially in harder scenarios.


QMIX(SEG) bears similarity with RODE~\cite{rode}. In particular, the procedures for clustering actions are the same. 
Just as  QMIX(SEG), RODE achieves semantic exploration by exploring at the higher level the space of roles. By the novel SEG exploration strategy, we recreate semantic exploration from RODE in QMIX. QMIX(SEG) shows in the MARL setting that exploration benefits in hierarchical algorithms (such as RODE) can be attained in non-hierarchical agents with certain exploration techniques, allowing non-hierarchical agents to achieve competitive results with hierarchical agents.

In addition to capturing the exploration benefits of RODE, QMIX(SEG) removes a restriction in RODE.
RODE requires that after a role is selected, an agent follows the selected role for $c$ time steps, that is, for $c$ time steps it will only take actions in the restricted action space corresponding to the role, even during test time.\footnote{As a consequence, RODE removes roles with only a single action and adds these actions to each of the remaining roles. This is because being assigned a singleton role forces an agent to take the same action for $c$ steps, which is undesirable. The removal of the singleton groups is not necessary in QMIX(SEG).} 
This greatly reduces the flexibility and level of freedom with which agents can select actions, and it can be disastrous for an agent that is unable to change its role promptly in response to some rapid changes in the environment. 
This restriction is removed in QMIX(SEG).

QMIX(SEG) further differs from RODE in three aspects. First, RODE focuses on efficiently learning a set of roles to decompose complex tasks while QMIX(SEG) focuses on improving the exploration for MARL. Second, RODE achieves improved exploration through its bi-level hierarchical agents while QMIX(SEG)  makes use of non-hierarchical agents and employs hierarchy only in action selection, which makes QMIX(SEG) computationally less expensive. 
Third, RODE uses Q-values of roles to assign roles to agents while QMIX(SEG) assigns groups to agents uniformly randomly.


\section{Experiments}

In our experiments, we have two goals. First, we would like to provide empirical evidence for our discussion in Section 3.4. We mentioned that in scenarios where advantageous states can only be reached by jointly selecting actions in groups containing a small number of actions, using SEG for exploration gives rise to better performance than using $\varepsilon$-greedy for exploration. We prove this by comparing SEG and $\varepsilon$-greedy on a simple coordination game we designed. Second, we would like to find out whether improved exploration with SEG leads to better performance in complicated scenarios. We study this problem by performing experiments on the popular SMAC benchmark~\cite{smac}.

\subsection{Coordination Game}

\begin{table*}[t]
    \centering
    \begin{tabular}{c c r r r r r r r}
         \specialrule{1.5pt}{1pt}{1pt}
         Map & Map Difficulty & QMIX & RODE & MAPPO(FP) & MAPPO(AS) & DDN & QMIX(SEG)\\
         \specialrule{1pt}{1pt}{1pt}
          MMM2 & Super Hard & 82.8(4.0) & 89.8(6.7) & 51.6(21.9) & 28.1(29.6) & \textbf{97.2} & 96.1(2.3)\\
          3s5z\_vs\_3s6z &  & 56.2(11.3) & 96.8(25.11) & 75.0(36.3) & 18.8(37.4) & 94.0 & \textbf{99.2(1.8)}\\
          27m\_vs\_30m &  & 34.4(5.4) & \textbf{96.8(1.5)} & 93.8(3.8) & 89.1(6.5) & 91.5 & 89.8(5.3)\\
          6h\_vs\_8z &  & 3.1(1.5) & 78.1(37.0) & 78.1(5.6) & 81.2(31.8) & \textbf{83.9} & 82.8(4.9)\\
          corridor &  & 64.1(14.3) & 65.6(32.1) & 93.8(3.5) & 93.8(2.8) & \textbf{95.4} & 93.8(23.5)\\
          \specialrule{1.5pt}{1pt}{1pt}
    \end{tabular}
    \caption{Median evaluation win rate and standard deviation on 5 super hard maps. Numbers for QMIX, RODE, MAPPO(FP), and MAPPO(AS) are obtained from the MAPPO paper~\protect\cite{mappo}. Numbers for DDN are obtained from the DFAC framework paper~\protect\cite{dfac}. The highest win rates for each map are bolded. For each map, each algorithm is trained for the same number of time steps as RODE.}
    \label{tab:data}
\end{table*}

To find out whether SEG allows agents to learn better strategy than $\varepsilon$-greedy in scenarios where advantageous states are hard to reach, we compare SEG and $\varepsilon$-greedy on a proof-of-concept coordination game. The game is specified by a 3-tuple $(N, K, M)$, where $N$ is the number of agents, $K$ is the number of steps agents must coordinate to reach the advantageous state, and $M$ is the number of actions in the action space. The states of the game are $\{s_0,\dots,s_{K}\}$, where $s_K$ is the advantageous state that agents should cooperate to visit. All $N$ agents start at state $s_0$ and have action space $A=\{a_0,\dots,a_{M-1}\}$. There is only one way to receive a positive reward in this game, which is to reach $s_K$. In $s_K$, any joint action selected gives a reward of 100 and causes the state to transition to $s_0$. Any joint action selected in any state other than $s_K$ gives a reward of 0. The only way to reach $s_{K}$ is to repeatedly take the joint action $(a_0,\dots,a_0)$, which causes the state to transition from $s_{j}$ to $s_{j+1}$ for $0\leq j<K$. Taking any joint action other than $(a_0,\dots,a_0)$ in $s_j$ for $0\leq j<K$ causes the state to transition to $s_0$.

\begin{figure}[t]
\centering
\subfigure[]{\label{fig:NK}\includegraphics[width=40mm]{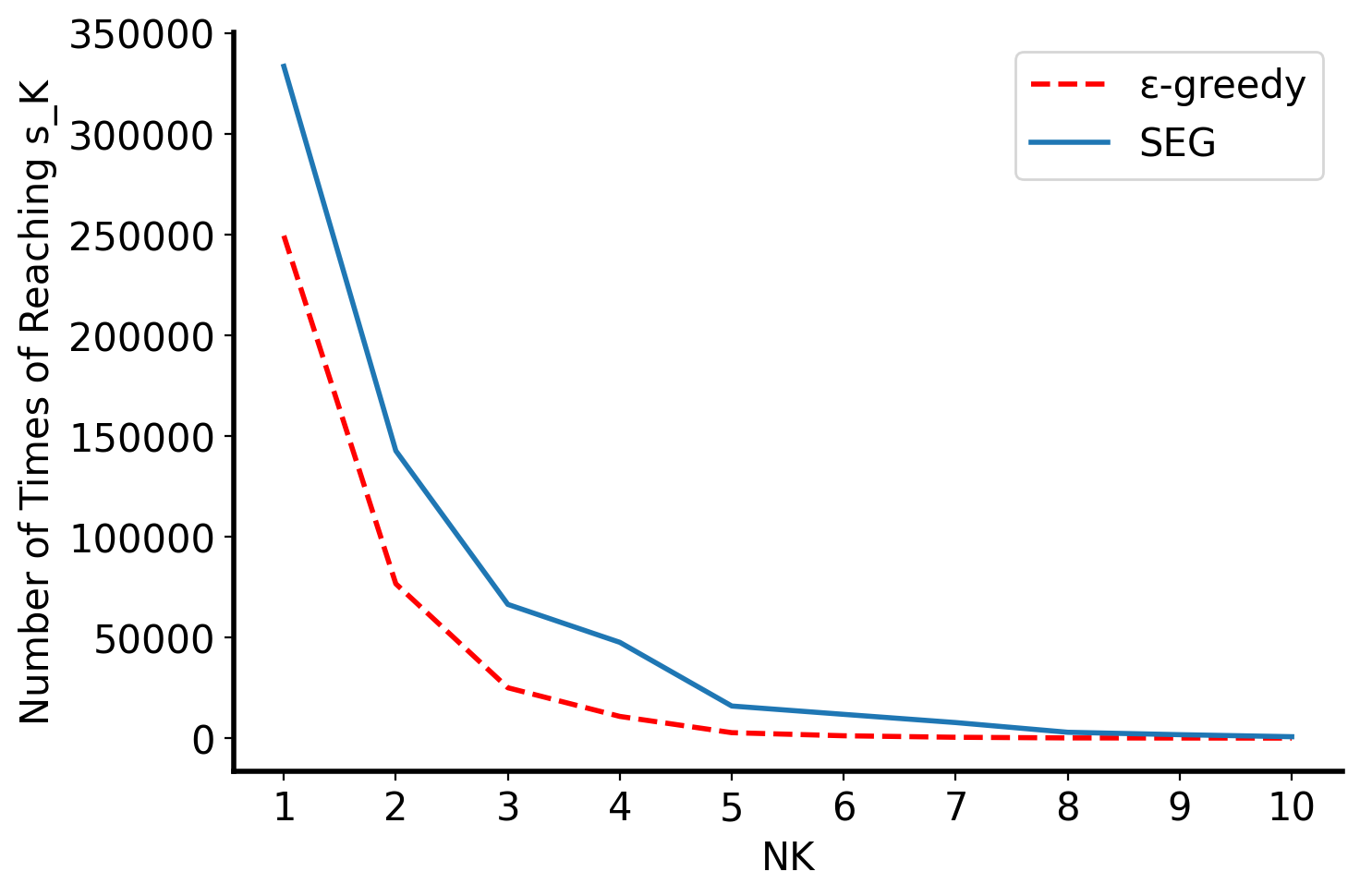}}
\hfill
\subfigure[]{\label{fig:Learn}\includegraphics[width=40mm]{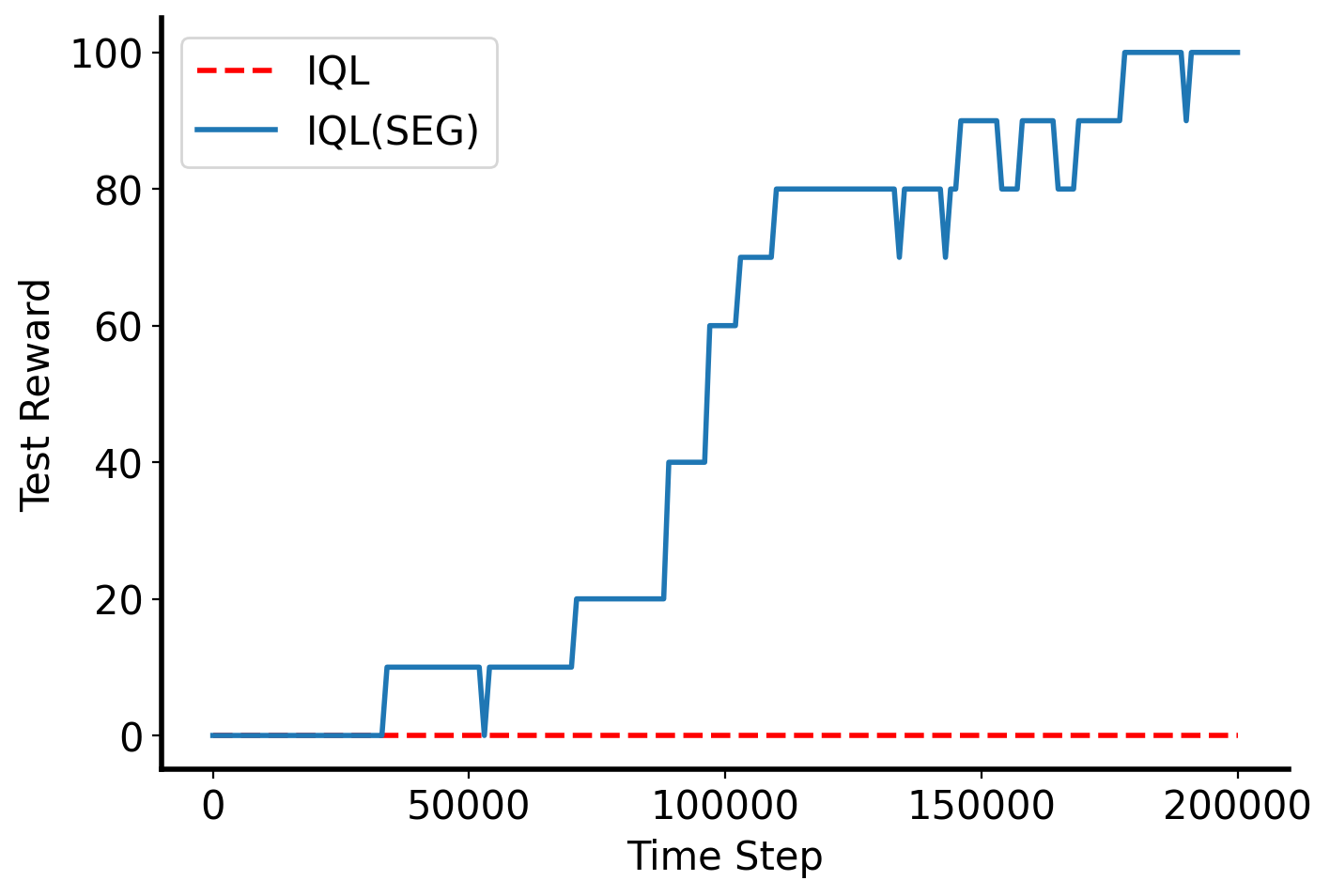}}
\caption{(a) The number of times $s_K$ is reached in one million steps as $NK$ increases. We choose $M=3$ and $\varepsilon=0.5$. (b) Test reward of IQL(SEG) and IQL on the coordination game with $N=5$, $K=4$, and $M=3$.}
\label{fig:CoordinationGame}
\end{figure}

The game poses a difficult challenge to agents. To receive the reward, the $N$ agents must coordinate and all select $a_0$ for consecutive $K$ steps to reach $s_K$. The parameters $N$ and $K$ control how difficult it is to reach $s_K$. When $N$ increases, more agents are required to coordinate. When $K$ increases, agents are required to coordinate for a larger number of time steps. We initialize $Q_i(s,a_0)=0\ \forall i,s$ and $Q_i(s, a_j)=0.1\ \forall i, s, j\neq 0$ such that the agents do not select $(a_0,\dots,a_0)$ at the beginning. The agents must reach $s_K$ via exploration to learn the optimal strategy of visiting $s_K$.

The number of available actions $M$ determines how difficult it is for an agent using $\varepsilon$-greedy for exploration to select $a_0$. Since $\varepsilon$-greedy exploration explores each action with equal probability, when $M$ increases, $\varepsilon$-greedy exploration has a lower probability of exploring $a_0$. When $M>2$, $\varepsilon$-greedy is biased towards selecting one of $a_1,\dots,a_{M-1}$ and it is difficult for agents using $\varepsilon$-greedy for exploration to reach $s_K$. On the other hand, using SEG for exploration does not have this problem. In this game, we choose the number of groups to be 2. Since $a_1, \dots, a_{M-1}$ have the same effects, the actions are split into two groups: $A^1=\{a_0\}$ and $A^2=\{a_1,\dots,a_{M-1}\}$. Since SEG first selects a group uniformly randomly before selecting an action, using SEG for exploration gives the same probability of selecting $a_0$ and selecting one of $a_1,\dots,a_{M-1}$, thus giving a higher probability of visiting state $s_K$. To see whether this is true empirically, we count the number of times of visiting $s_K$ by exploration for SEG and $\varepsilon$-greedy. Since we are interested in the probability $s_K$ is visited {\it by exploration}, we initialize the Q-values as mentioned and do not update the Q-values. We present the number of times of visiting $s_K$ by exploration for SEG and $\varepsilon$-greedy in one million steps as $NK$ increases in Figure \ref{fig:NK}.

To see whether using SEG for exploration allows agents to learn a better strategy on this game, we train IQL~\cite{iql} using SEG for exploration (IQL(SEG)) and IQL using $\varepsilon$-greedy for exploration (IQL) on this game with $N=5$, $K=4$ and $M=3$. We anneal $\varepsilon$ from $1.0$ to $0.05$ in $30000$ steps. We run 10 trials and in each trial, we evaluate the algorithms for $K+1$ steps on a regular interval. We present the mean reward of the 10 trials in Figure \ref{fig:Learn}.

From Figure \ref{fig:NK}, it can be observed that SEG allows the agents to explore $s_K$ for a larger number of times than $\varepsilon$-greedy. When $NK=10$, using SEG for exploration still allows the agents to reach $s_K$ for around 700 times while using $\varepsilon$-greedy only allows the agents to reach $s_K$ for less than $10$ times. This proves that SEG, by removing the bias in $\varepsilon$-greedy, allows agents to visit advantageous states more when such states can only be reached by a small number of similar actions. From Figure \ref{fig:Learn}, it can be observed that SEG, by allowing agents to visit the advantageous state $s_K$ for a much larger number of times, allows the IQL algorithm to learn the optimal behavior of reaching $s_K$, getting a reward of 100. On the other hand, using $\varepsilon$-greedy for exploration fails to give rise to the optimal behavior. Therefore, in scenarios where agents have to jointly select from a small number of actions with similar effects for extended time steps to reach advantageous states, SEG allows agents to better reach these states and thus allows the agents to learn better strategies.

\subsection{StarCraft Multi-Agent Challenge}

To find out whether improved exploration with SEG can lead to better performance in complicated scenarios, we evaluate our method QMIX(SEG) on SMAC~\cite{smac}. We choose SMAC since it allows algorithms to be compared directly by their win rates on different maps. In SMAC, maps can be classified into three difficulties: easy, hard, and super hard. Hard maps pose different challenges such as handling large action space and kiting. Super hard maps are hard-exploration problems, making them appropriate scenarios to test our method.

We compare QMIX(SEG) with QMIX to directly see whether the improved exploration over $\varepsilon$-greedy can provide better performance. We also compare QMIX(SEG) with RODE~\cite{rode} to see whether QMIX(SEG), capturing the exploration benefits and avoiding the limitations of RODE, can outperform RODE. We further compare QMIX(SEG) with state-of-the-art algorithms on SMAC: MAPPO~\cite{mappo} and DDN~\cite{dfac}. We follow the evaluation procedure outlined in the MAPPO paper~\cite{mappo}. 

We present the results of QMIX(SEG) and baselines on all 5 super hard maps in Table~\ref{tab:data}.  
QMIX(SEG) outperforms most state-of-the-art algorithms, including QMIX, RODE, and MAPPO. Compared with MAPPO(FP) and MAPPO(AS), QMIX(SEG) achieves higher win rates on 2 maps and similar win rates on the remaining 3 maps. DDN makes use of distributional reinforcement learning and is much more complex than our approach. Still, QMIX(SEG) is able to obtain a higher win rate than DDN on 1 map and obtain win rates within 2\% of that of DDN on all remaining maps.

By comparing QMIX(SEG) and QMIX, it can be observed that using SEG instead of $\varepsilon$-greedy for exploration significantly improves the performance on all maps. On harder maps, agents are usually required to jointly move for extended time steps to reach advantageous positions on the map. Also, movement actions are outnumbered by attack actions. SEG, by removing the bias in $\varepsilon$-greedy towards selecting attack actions, allows agents to jointly explore movement actions more and reach advantageous positions with higher probability, leading to better performance on harder maps. 

QMIX(SEG) has higher win rates than RODE on 4 out of 5 super hard maps. Removing the limitations of RODE may be the reason QMIX(SEG) can achieve better performance on these maps. A difference between QMIX(SEG) and RODE is that RODE uses the Q-values of roles to determine which role to assign to an agent while we assign groups to agents uniformly randomly with probability $\varepsilon$. From the close performance of QMIX(SEG) and RODE, it can be concluded that how we assign roles or groups, and correspondingly action spaces, to agents do not matter. We just need to make sure that the action spaces of agents are restricted with a certain probability to gain the exploration benefits.

\section{Conclusion}
We propose a novel non-hierarchical learning approach QMIX(SEG), which utilizes QMIX for value function factorization, and a novel hierarchical dithering exploration method SEG. SEG is a simple, general, and effective way to improve exploration by semantic exploration in the space of groups of similar actions and removing the bias in $\varepsilon$-greedy. From the experiments, we show that SEG significantly outperforms $\varepsilon$-greedy, especially in hard scenarios. QMIX(SEG) can achieve state-of-the-art performance on SMAC. Our work also provides yet another compelling example of extracting exploration methods from hierarchical reinforcement learning algorithms. Future work can focus on combining our work with temporally extended $\varepsilon$-greedy and applying SEG on single agent reinforcement learning problems.

$\ $

\bibliographystyle{named}
\bibliography{22}

\begin{table*}[t]
    \centering
    \begin{tabular}{c c r r r r r r r}
         \specialrule{1.5pt}{1pt}{1pt}
         Map & Map Difficulty & QMIX & RODE & MAPPO(FP) & MAPPO(AS) & QMIX(SEG)\\
         \specialrule{1pt}{1pt}{1pt}
          2s\_vs\_1sc & Easy & 96.9(1.2) & \textbf{100.0(0.0)} & \textbf{100.0(0.0)} & \textbf{100.0(0.0)} & \textbf{100.0(1.2)}\\
          2s3z  &  & 95.3(3.9) & \textbf{100.0(0.0)} & 96.9(1.5) & 96.9(1.5) & \textbf{100.0(1.2)}\\
          3s5z &  & 85.9(4.6) & 93.75(2.0) & 71.9(11.8) & 53.1(15.4) & \textbf{97.7(1.9)}\\
          1c3s5z &  & 95.3(1.2) & \textbf{100.0(0.0)} & \textbf{100.0(0.0)} & 96.9(2.6) & \textbf{100.0(0.6)}\\
          10m\_vs\_11m &  & 82.8(4.1) & 95.3(2.2) & 81.2(8.3) & 89.1(5.5) & \textbf{98.4(1.4)}\\
          2c\_vs\_64zg & Hard & 70.3(3.8) & \textbf{100.0(0.0)} & 96.9(3.1) & 95.3(3.5) & 96.9(1.5)\\
          bane\_vs\_bane &  & \textbf{100.0(0.0)} & \textbf{100.0(46.4)} & \textbf{100.0(0.0)} & \textbf{100.0(0.0)} & \textbf{100.0(5.2)}\\
          5m\_vs\_6m &  & 54.7(3.5) & 71.1(9.2) & 65.6(14.1) & 68.8(8.2) & \textbf{94.5(2.2)}\\
          3s\_vs\_5z &  & 56.2(8.8) & 78.9(4.2) & 98.4(5.5) & \textbf{100.0(1.2)} & 36.7(29.6)\\
          \specialrule{1.5pt}{1pt}{1pt}
    \end{tabular}
    \caption{Median evaluation win rate and standard deviation on 5 easy and 4 hard maps. Numbers for QMIX, RODE, MAPPO(FP), and MAPPO(AS) are obtained from the MAPPO paper~\protect\cite{mappo}. The highest win rates for each map are bolded. For each map, each algorithm is trained for the same number of time steps as RODE.}
    \label{tab:data2}
\end{table*}

\begin{table}[t]
  \centering
  \begin{tabular}{c r}
  \specialrule{1.5pt}{1pt}{1pt}
    Map & $\varepsilon$ anneal time (Thousand Steps) \\
    \specialrule{1pt}{1pt}{1pt}
    2s\_vs\_1sc & 50\\
    2s3z & 50 \\
    3s5z & 50 \\
    1c3s5z & 50 \\
    10m\_vs\_11m & 50 \\
    2c\_vs\_64zg & 50 \\
    bane\_vs\_bane & 50 \\
    5m\_vs\_6m & 50 \\
    3s\_vs\_5z & 300 \\
    MMM2 & 50 \\
    3s5z\_vs\_3s6z & 500\\
    27m\_vs\_30m & 50 \\
    6h\_vs\_8z & 300\\
    corridor & 500 \\
  \specialrule{1.5pt}{1pt}{1pt}
  \end{tabular}
  \caption{$\varepsilon$ anneal time of QMIX(SEG) on different maps.}
  \label{tab:annealTime}
\end{table}

\section*{A\quad Hyperparameters}

For learning action representations, we choose the dimension of action representations $d$ to be 20 and $\lambda_e=10$. Following RODE, we train the encoder $f_e$ and the predictors $p_o$, and $p_r$ in only the first $50K$ steps. It is shown in the RODE paper~\cite{rode} that $50K$ steps are enough to learn an action encoder that transforms a one-hot action to a representation vector that captures the effect of the action on the environment and other agents. For $k$-means clustering, same as RODE, we choose the number of clusters to be 3 for homogeneous enemies, 5 for heterogeneous enemies, and 2 for single enemy.

The architecture of the Q-network and the mixing network in QMIX(SEG) follows that of QMIX~\cite{qmix} and are trained using Adam with learning rate 5e-4, $\beta_1=0.9$, $\beta_2=0.99$, $\epsilon=$1e-5, and with no weight decay. The implementation of QMIX(SEG) is modified from the implementation of QMIX by the Whiteson Research Lab and does not include the implementation tricks in the MAPPO paper. We expect the performance of QMIX(SEG) to further improve if these tricks are implemented. The batch size used is 32. All experiments on the SMAC benchmark follow the default settings~\cite{smac}. 

The $\varepsilon$ for SEG is the same across two levels and annealed from 1 to 0.05 over different time steps for different maps, ranging from $50K$ steps to $500K$ steps. We find that the $\varepsilon$ anneal time is an important hyperparameter affecting the performance of QMIX(SEG) on harder maps. We present the $\varepsilon$ anneal time of QMIX(SEG) on different maps used in our experiments in Table~\ref{tab:annealTime}.

\section*{B\quad Additional Experiment Results}

The performances of QMIX(SEG) and baselines on 5 easy maps and 4 hard maps of SMAC are presented in Table~\ref{tab:data2}. QMIX(SEG) can achieve the highest win rates on all of the easy maps. QMIX(SEG) achieves the highest win rates on 2 hard maps and a win rate close to the highest on another hard map. QMIX(SEG) outperforms QMIX, MAPPO(FP), and MAPPO(AS) on easy and hard maps, obtaining similar or higher win rates than these baselines on all but one map. QMIX(SEG) has close performance as RODE on easy and hard maps.

The only map where QMIX(SEG) struggles to obtain good performance is \textbf{3s\_vs\_5z}. On this map, agents are required to alternate between attacking for one step and moving for several steps. Since QMIX(SEG) does not repeat a group for multiple steps as in RODE, QMIX(SEG) requires more time than RODE to learn to move for multiple steps after attacking once. When we extend the number of steps to train to $5M$, QMIX(SEG) achieves a median win rate of 98.4\% with standard deviation 8.92\% on \textbf{3s\_vs\_5z}.

\end{document}